\documentclass{article}
\usepackage{spconf,amsmath,graphicx}
\usepackage{subfigure}

\usepackage{enumitem}
\usepackage{wrapfig}
\usepackage{caption}
\usepackage{epstopdf}

\title{Disjunctive Normal Level Set: An Efficient Parametric Implicit Method}

\name{Fitsum Mesadi $^{\star}$ \sthanks{supported by NIH 1R01-GM098151-01 and  NSF IIS-1149299(TT)} \qquad Mujdat Cetin $^{\dagger}$ \sthanks{supported by TUBITAK-113E603 and TUBITAK-2221 grants} \qquad Tolga Tasdizen $^{\star}$ }
 \address{$^{\star}$ Department of Electrical and Computer Eng., University of Utah, United States \\ $^{\dagger}$ Faculty of Eng. and Natural Sciences, Sabanci University, Turkey}
\begin{document}
%\ninept
%
\maketitle
\begin{abstract}
Level set methods are widely used for image segmentation because of their capability to handle topological changes. 
%However, in spite of the numerous works in the literature, level set methods still have several drawbacks: irregularities of the signed distance function during evolution, slow computational speed, and the dramatic increase in the computational cost and memory requirement as the number of objects to be simultaneously segmented grows. 
In this paper, we propose a novel parametric level set method called Disjunctive Normal Level Set (DNLS), and apply it to both two phase (single object) and multiphase (multi-object) image segmentations. The DNLS is formed by union of polytopes which themselves are formed by intersections of half-spaces. The proposed level set framework has the following major advantages compared to other level set methods available in the literature. First, segmentation using DNLS converges much faster. Second, the DNLS level set function remains regular throughout its evolution. Third, the proposed multiphase version of the DNLS is less sensitive to initialization, and its computational cost and memory requirement remains almost constant as the number of objects to be simultaneously segmented grows. 
The experimental results show the potential of the proposed method. 
\end{abstract}
\section{Introduction}
\label{sec:intro}
Level set method, first introduced by Osher and Sethian \cite{Osher1988}, is a popular technique for following the evolution of interfaces. The technique has a wide range of applications in image processing, computer graphics, computational geometry, optimization, and computational fluid dynamics. 

\noindent
{\bf Related Work:}
The ability of the level set method to handle topological changes has made the algorithm suitable for image segmentation and tracking applications. However, the traditional level set formulation has several major drawbacks. Next, we briefly discuss some of these limitations and the related work in the literature to address them. 
 
The original level set method is computationally expensive and the level set function can also develop irregularities (such as very sharp or flat shape) during evolution \cite{Li2005}. The large computational cost is mainly because the level set method increases the dimension of the problem by one, and the gradient descent methods used in level set evolution require large number of iterations, since the time step is limited by the standard Courant-Friedrichs-Lewy (CFL) condition \cite{Aghasi2011}. Modifications such as fast marching, sparse, and narrow band schemes were proposed in order to improve the computational time of level set evolution; however, level set method is still a relatively “slow” approach \cite{Duan2010}. In order to overcome the irregularities of the signed distance function, some of the techniques employed are periodic re-initializing of the level set function, and adding a regularizing term that forces the level set function to be close to a signed distance function \cite{Li2010, Zhang2013}. However, the re-initialization method has the undesirable effects of moving the level set from its original location, expensive computational cost, blocking the emerging of new contours; whereas, the regularizing terms still cannot guarantee the smoothness of the signed distance functions \cite{Li2010, Zhang2013}. 

In order to limit the computational cost and also minimize irregularities of the level set function, recently parametric level set methods are proposed in the literature \cite{Gelas2007, Bernard2007, Bernard2009, Aghasi2011, Luo2007, Pingen2010}. Parametric methods have faster computational speed since they reduce the dimensionality of the problem. Although, the parametric methods in the literature simplify the challenges involved in keeping the regularities of the level set function, they still require re-normalization of the level-set function during the evolution process \cite{Bernard2009}.  

Furthermore, the techniques available in the literature on multiphase level set formulations for simultaneous segmentation of multiple objects have the drawback of large increase in computational time and memory requirement as the number of objects to be segmented increases. Most level set based multi-object segmentation approaches available in the literature use $R$ level set functions to segment $R$ objects \cite{Zhao1996, Brox2006}. In \cite{ChanVeseMultiphase} Vese and Chan proposed a multiphase level set framework that requires $\log R$ level set functions to segment $R$ objects. Although \cite{ChanVeseMultiphase} is relatively computationally attractive, in applications that require tracking of the individual objects and use of their shape priors, it is more convenient to have a unique level set for each object. In addition, most multiphase level set methods in the literature are very sensitive to the initialization. The level set method we propose in this paper has the highly desirable properties that it is less sensitive to initialization, and its computational cost and memory requirement remains almost constant as the number of objects to be segmented grows, while also having the capability to represent each object with unique level set. 

\noindent
{\bf Contributions:}
In this paper, we propose a novel parametric level set method called Disjunctive Normal Level Set (DNLS), and apply it to both two phase and multiphase image segmentations. The DNLS is based on an implicit and parametric shape model called Disjunctive Normal Shape Models (DNSM). The DNSM is recently used for a single object segmentation: to model the shape and appearance priors of objects in ~\cite{Mesadi2015, Usman2016}, and as an interactive segmentation framework in~\cite{Ramesh2015}. The DNLS we propose keeps the level set function regular all the time, and it has much faster computational speed since its time step is not limited by the standard CFL. This paper has three major contributions. First, we present the DNLS level set representation (in Section \ref{sec:DNFlevelSet}). Second, we give a two-phase region-based segmentation using the proposed DNLS method (in Section \ref{sec:RegionBased}). Third, we present a novel DNLS multiphase segmentation (in Section \ref{sec:MultiPhase}) which is robust to initialization and efficient in terns of both the computational time and memory requirement.  		

\section{Method}
\label{sec:Method}
In this section, first we present the disjunctive normal form level set representation, and then we show how the proposed level set can be used for region-based image segmentation in both two-phase and multiphase cases. 
  \vspace{-0.2cm}

\subsection{Disjunctive Normal Level Set Representation}
\label{sec:DNFlevelSet}
\begin{wrapfigure}{R}{0.14\textwidth}
 \vspace{-0.7cm}
		\begin{center}
       \subfigure[]{%
            \includegraphics[width=0.06\textwidth]{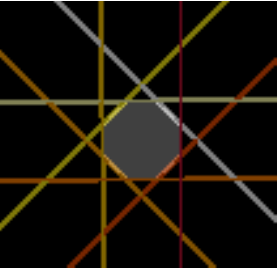}
        }%
 				\subfigure[]{%
           \includegraphics[width=0.07\textwidth]{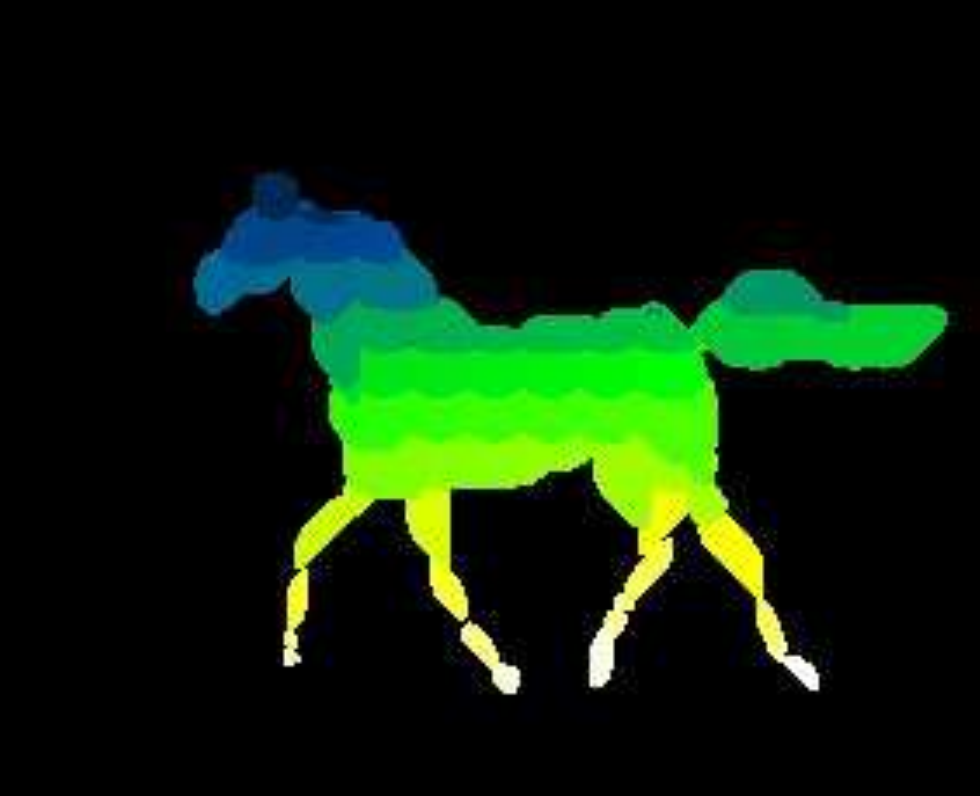}
        }%										
   \end{center}
   \vspace{-0.7cm}
    \caption{%
   % DNLS regions 
     }%
   \label{fig:DNSMIllustration}
	 \vspace{-0.3cm}
\end{wrapfigure}
We begin the description of the disjunctive normal level set shape representation with an example for clarity. Figure~\ref{fig:DNSMIllustration}(a) shows how the conjunctions of eight half-spaces forms a convex polytope. Our DNLS uses the disjunction of convex polytopes to represent complex shapes, as shown in Fig.~\ref{fig:DNSMIllustration}(b). 

Consider the characteristic function $f:\mathbf{R}^D\rightarrow\mathbf{B}$ where $\mathbf{B}=\{0,1\}$.  Let $\Omega^+=\{\mathbf{x}\in \mathbf{R}^D:f(\mathbf{x})=1\}$. Let us approximate $\Omega^+$ as the union of $N$ convex polytopes $\tilde{\Omega}^+=\cup_{i=1}^{N} {\cal P}_i$, where the {\em i}'th polytope is defined as the intersection of ${\cal P}_i=\cap_{j=1}^{M} {\cal H}_{ij}$ of $M$ half-spaces. $H_{ij}$ is defined in terms of its indicator function 
\begin{equation}
h_{ij}(\mathbf{x})=\left\{ 
\begin{array}{lr}
1,& \sum_{k=0}^D w_{ijk}x_k+b_{ij}\geq 0\\
0,& {otherwise}
\end{array}
\right.,
\label{eq:h}
\end{equation}
where $w_{ijk}$ and $b_{ij}$ are the weights and the bias term, and $D$ is the dimension. Since any Boolean function can be written in disjunctive normal form~\cite{hazewinkel1997}, we can construct $\tilde{f}(\mathbf{x}) = \bigvee_{i=1}^N{ \underbrace{\left(\bigwedge_{j=1}^{M} h_{ij}(\mathbf{x})\right)}_{B_i(\mathbf{x})} } $
such that $\tilde{\Omega}^+=\{\mathbf{x}\in \mathbf{R}^n:\tilde{f}(\mathbf{x})=1\}$. Since $\tilde{\Omega}^+$ is an approximation to $\Omega^+$, it follows that $\tilde{f}$ is an approximation to $f$. Our next step is to provide a differentiable approximation to $\tilde{f}$, which is important because it allows us use variational approaches; in other words, it allows us to formulate various energy functions and to minimize them with respect to the parameters of the model. First, the conjunction of binary variables $ \bigwedge_{j=1}^{M} h_{ij}(\mathbf{x}) $ can be replaced by the product $\prod_{j=1}^M h_{ij}(\mathbf{x})$. Then, using De Morgan's laws~\cite{hazewinkel1997} we replace the disjunction of the binary variables $\bigvee_{i=1}^N B_{i}(\mathbf{x}) $ with $\lnot\bigwedge_{i=1}^N\lnot{B_i(\mathbf{x})}$, which in turn can be replaced by the expression $1 - \prod_{i=1}^N (1- B_{i}(\mathbf{x}))$. Finally, we approximate $h_{ij}(\mathbf{x})$ with logistic sigmoid functions $\sigma_{ij}(\mathbf{x}) = \frac{1}{1+e^{\sum_{k=0}^D w_{ijk}x_k+b_{ij}}}$ to get the differentiable approximation of the characteristic function $\hat{f}(\mathbf{x})$
 \vspace{-0.7cm}
\begin{equation}
\hat{f}(\mathbf{x};\mathbf{W}) = 1 - \prod _{\substack{i=1 \\ i \in \aleph(\mathbf{x})}}^N 
\left(1-  
\underbrace{
\prod_{j=1}^M 
\frac{1}{1+e^{\sum_{k=0}^D w_{ijk}x_k+b_{ij}}}
}_{g_i(\mathbf{x})}
\right),
\label{eq:DNLS}
\end{equation}
where $\aleph (\mathbf{x})$ is the list of polytopes that are in the neighborhood of the location $\mathbf{x}$.
\begin{wrapfigure}{R}{0.07\textwidth}
	 \vspace{-0.7cm}
		\begin{center}
%       
        %\subfigure[]{%
           \includegraphics[width=0.065\textwidth]{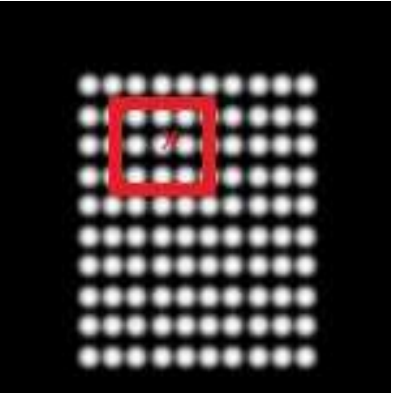}
        %}%											
%
   \end{center}
   \vspace{-0.5cm}
    \caption{%
   % DNLS regions 
     }%
   \label{fig:DNLS_Initialization}
	 \vspace{-0.3cm}
\end{wrapfigure}

The DNLS formulation of equation (\ref{eq:DNLS}) is similar to the DNSM shape model presented in ~\cite{Mesadi2015, Ramesh2015}, except for two modifications. First, instead of using the application domain knowledge to decide on the small number of polytopes needed ~\cite{Mesadi2015, Ramesh2015} , we use a large number of polytopes, $N$, in the DNLS formulation, and initialize the level set with regularly distributed polytopes (in the region of interest), as can be seen in Fig.~\ref{fig:DNLS_Initialization}. The use of dense initialization helps the DNLS to capture complex shapes, detect small parts and holes, and provides a fast convergence speed. Second, for computational efficiency, we only use the neighboring polytopes, $\aleph (\mathbf{x})$, for each location, $\mathbf{x}$, in the image. For instance, in Fig.~\ref{fig:DNLS_Initialization} only the polytopes in the red box are used when evaluating the characteristic function, $f$, at location $\mathbf{x}$.  
During the level set evaluation, the individual polytopes can grow, shrink, deform, disappear, and reappear. The only adaptive parameters in equation (\ref{eq:DNLS}) are the weights and biases of the first layer of logistic sigmoid functions $\sigma_{ij}(\mathbf{x})$ which define the orientations and positions of the linear discriminants that form the shape boundary. In equation (\ref{eq:DNLS}) the level set $f (x) = 0.5$ is taken to represent the interface between the foreground $f(\mathbf{x})>0.5$ and background $f(\mathbf{x})<0.5$ regions. 
%Notice that $f$ is not a signed distance function. 

\subsection{Two-phase (One Object) Segmentation}
\label{sec:RegionBased}
%Image segmentation using level set can be categorized as region-based and edge-based methods (papers here)~\cite{Bertelli2010, Zhang2010}. Segmentation of images with region-based level set has two major advantages compared to the edge-based method. First, the region based methods are less sensitive to noise and weak edges. Second, the region-based algorithms are less dependent on the initialization of the level sets ~\cite{Bertelli2010, Zhang2010}. 
Mumford and Shah in \cite{Mumford1989} presented the first variational approach to region-based segmentation where the image is approximated by piecewise smooth regions. In \cite{ChanVese2001} Chan and Vese (CV) proposed one of the most popular region-based level set segmentation by approximating the image into piecewise constant regions and evolved the level set in order to minimize the variance of each partition. The CV equivalent two-phase region-based variational energy using the proposed DNLS level set can be given as 
 \begin{equation} 
  %\begin{split}
 %E(\mathbf{W}) = \int_{\Omega} \left((I(\mathbf{x})-c_{1})^2 f(\mathbf{x}) + (I(\mathbf{x})-c_{2})^2 (1-f(\mathbf{x}))\right) d\mathbf{x}
 E(\mathbf{W}) = \int_{\Omega}(I(\mathbf{x})-c_{1})^2 f(\mathbf{x}) + (I(\mathbf{x})-c_{2})^2 (1-f(\mathbf{x})) d\mathbf{x}
\label{eq:dnsm_Region}
	%\end{split}
\end{equation} 
%where the discriminant vector parameter $\mathbf{W}$ is as defined in Section \ref{sec:DNFlevelSet}, and the level set function $f$ is as given in equation (\ref{eq:DNLS}). 
where $c_{1}$  and $c_{2}$ are the average intensities in the foreground and background regions. $I(\mathbf{x})$ is the intensity of the pixel at location $\mathbf{x}$.

The energy minimization implies computing the derivatives of equation (\ref{eq:dnsm_Region}) with respect to each discriminant parameters, $w_{ijk}$. During segmentation, the update to the discriminant weights, $w_{ijk}$, is obtained by minimizing the energy using gradient descent as
\begin{equation} 
 % \begin{split}
 \frac{\partial E}{\partial w_{ijk}} =  \left((I(\mathbf{x})-c_{1})^2 - (I(\mathbf{x})-c_{2})^2 \right) \frac{\partial f}{\partial w_{ijk}}
\label{eq:dnsm_Region_Der}
	%\end{split}
\end{equation} 
where $ \frac{\partial f}{\partial w_{ijk}} = -\left(\prod_{\substack{r \in \aleph \\ r \neq i}}(1- g_{r}(\mathbf{x})) \right) g_{i}(\mathbf{x}) (1- \sigma_{ij}(\mathbf{x})) x_{k}$ 
Therefore, during the level set evolution the discriminant parameters are updated on each iteration as 
$w_{ijk} \leftarrow w_{ijk} -\gamma \frac{\partial E}{\partial w_{ijk}} $, where $\gamma$ is the step-size. Since the evolution of the proposed parametric level set is not constrained by the standard CFL condition, we can easily choose large $\gamma$ at the beginning of the evolution and gradually decrease as the segmentation progresses. It is also important to notice that the level set function, $f$, remains regular during the evolution.

\subsection{Multiphase (Multi-Object) Segmentation}
\label{sec:MultiPhase}
%\textit{ Here I give how to formulate a multiphase version of our DNF level set, Give the energy equations, explain the effetc of each energy term,  the derivation, the update to the discriminants }  \\[5\baselineskip]
In this section, we extend the DNLS framework to segmentation of images with more than two regions. In order to concurrently segment R objects (regions), techniques in the literature usually use $R$ or $\log R$ level sets. Since the DNLS level set presented in Section (\ref{sec:DNFlevelSet}) is made up of union of many polytopes, the single level set function given in equation (\ref{eq:DNLS}) can be used to segment R regions. Each of the polytopes can individually be regraded as a level set function, and hence, they can be assigned to different regions (objects). Therefore, the movement of the surfaces between the different objects takes place in two ways: by the deformations of the polytopes, and by change in the labels of the polytopes. The DNLS multiphase energy can then be given as $E = E_{L} + E_{D}$; where $E_{L}$ is the energy for changing the labels of the polytopes, and $E_{D}$ is the energy for deforming the polytopes. 

\begin{figure}[ht!] 
		\begin{center}
			\subfigure[]{%
            \includegraphics[width=0.09\textwidth]{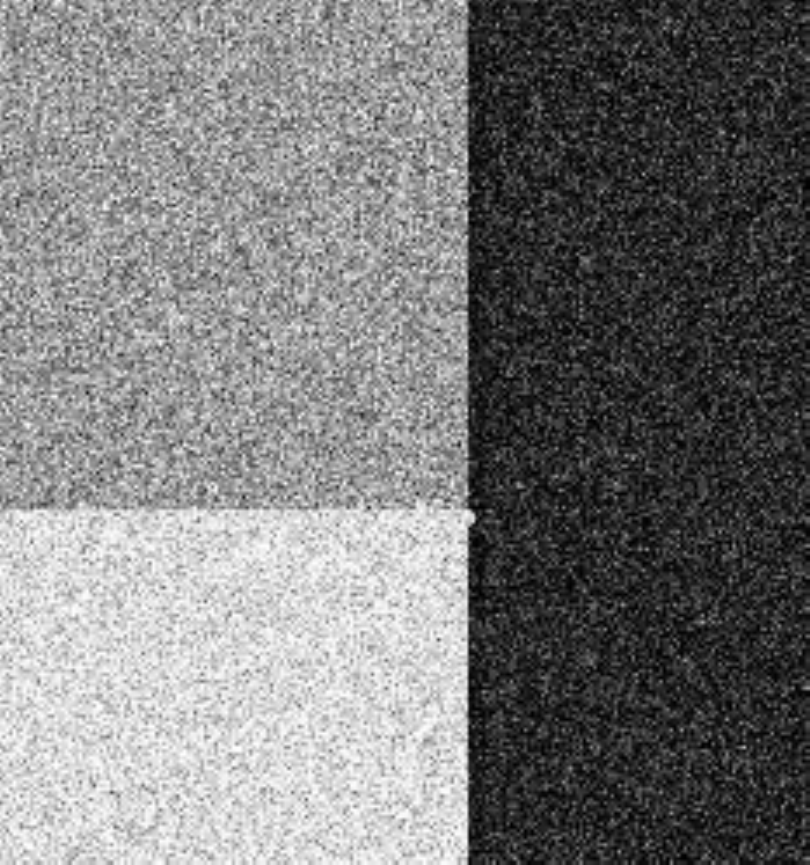}
        }%				
				\subfigure[]{%
            \includegraphics[width=0.09\textwidth]{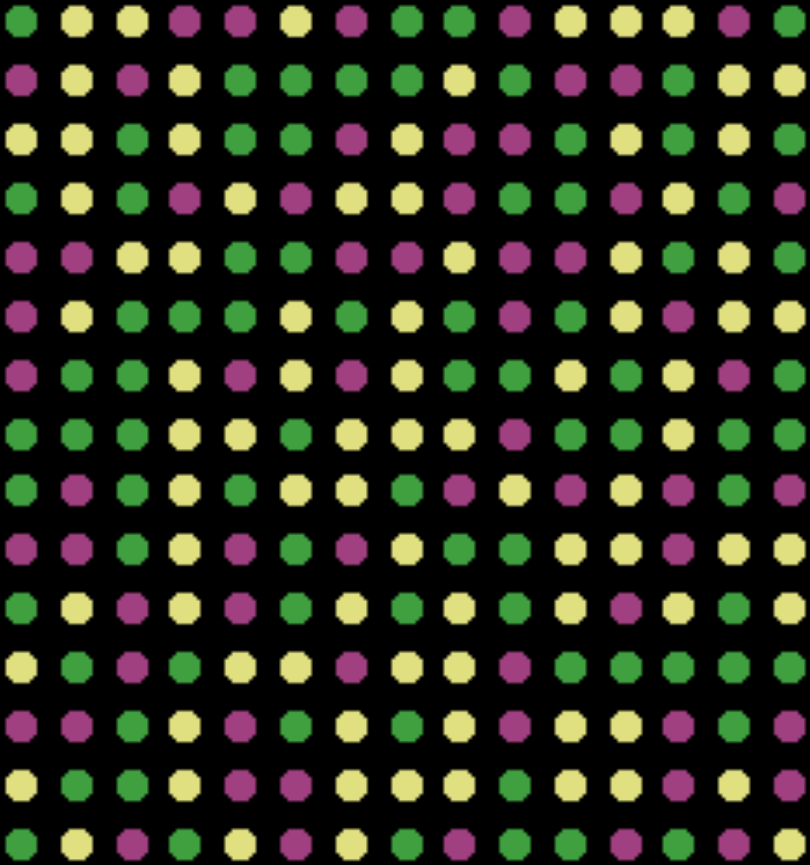}
        }%
 				\subfigure[]{%
           \includegraphics[width=0.09\textwidth]{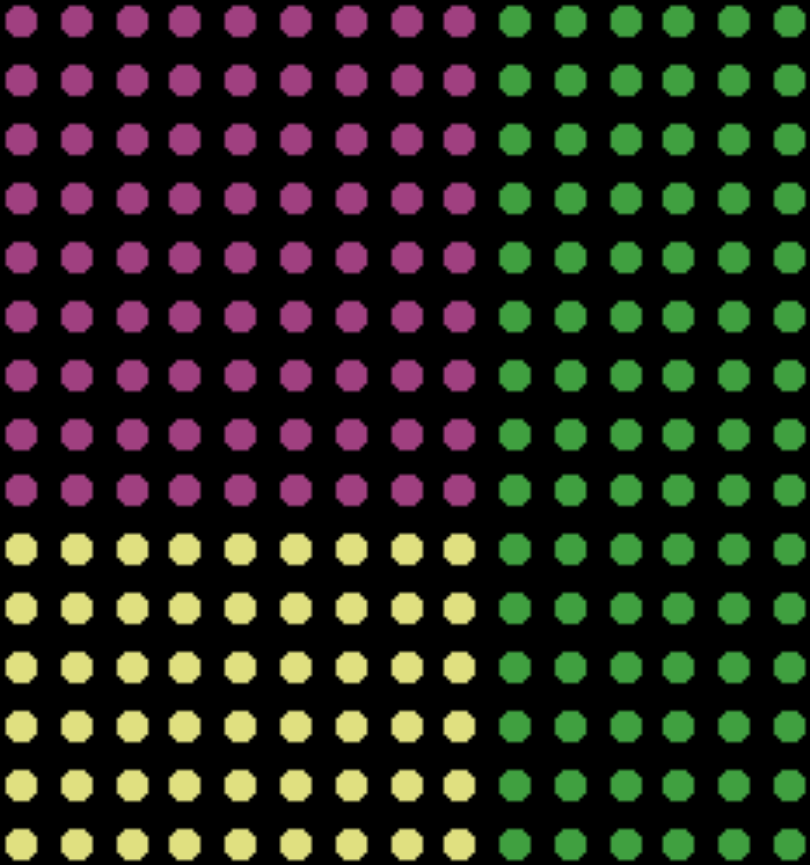}
        }%
				\subfigure[]{%
						 \includegraphics[width=0.0745\textwidth]{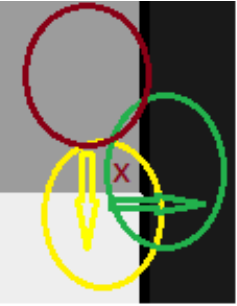}
        }%
				\subfigure[]{%
					            \includegraphics[width=0.09\textwidth]{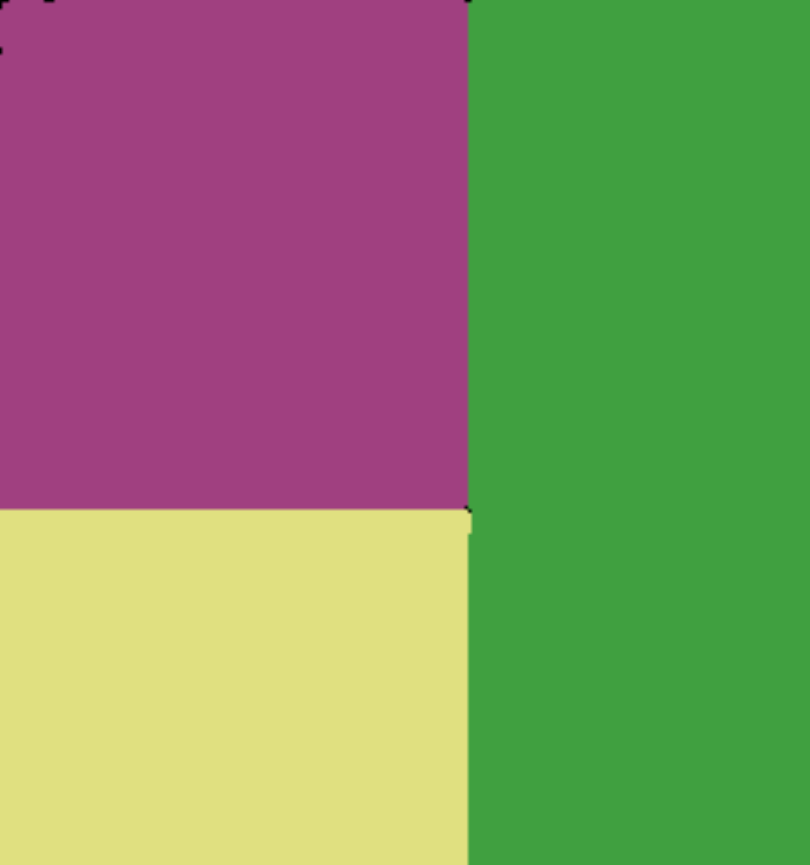}
        }%		
   \end{center}
   \vspace{-0.8cm}
    \caption{%
		a) Multiphase image. b) and c) show the effect of applying $E_{L}$. d) shows the effect of the $E_{D}$ term, using enlarged polytopes for clarity, and the arrows show the direction of their deformation. e) Final segmentation result.
     }%
   \label{fig:MultiDemo}
   \vspace{-0.3cm}
%\end{wrapfigure}
\end{figure}

For the label assigning energy term $E_{L}$ we use a simple K-means clustering; however, any advanced clustering method can be employed. By first computing the mean intensities for each polytope $p_i$, we can cluster them into R region labels. That is, given a set of N polytopes with their average intensities $(p_1, p_2, . . . , p_N)$, K-means clustering aims to partition the N polytopes into $R$ region label sets $L = \{L_1, L_2, . . . , L_R\}$ so as to minimize the within-cluster sum of squares as 
%\begin{equation} 
 $E_{L} = \operatornamewithlimits{argmin}_{L} \sum_{r=1}^{R} \sum_{p \in L_r} (p-c_r)^2$, 
%\label{eq:EnergyLabel}
%\end{equation} 
where $c_r$ is the mean of the intensity of all the polytopes assigned to region $r$ (the mean intensity of region $r$). Fig.~\ref{fig:MultiDemo}(b) and (c) show the effect of applying $E_{L}$, which changes the label assignments of the polytopes (the colors show the different labels).

\vspace{-0.3cm}
%\vspace{-0.3cm}

The deformation energy term, $E_{D}$, controls the deformation of the polytopes so as to achieve proper segmentation while at the same time avoiding the overlap of the different segments or creation of gaps, and is given by
 \vspace{-0.25cm}
\begin{equation} 
E_{D}(\mathbf{W}) =  \int_{\mathbf{x} \in \Omega} \left( -f_{rb} + \sum_{\substack{r=1 \\ r \neq rb; ~ r \in \aleph(\mathbf{x})}}^R f_r \right) \mathbf{dx}
\label{eq:DNLS_MultiED}
  \vspace{-0.25cm}
\end{equation}
where we represent each region, $r$, by a unique level set $f_r$, that is formed by the union of the polytopes with label $r$. The best level set, $f_{rb}$, for pixel location $\mathbf{x}$ with intensity $I(\mathbf{x})$ is the $r$ label that results in the smallest $(I(\mathbf{x})-c_r)^2$. That is, $f_{rb}$ is the best level set if, ~ 
$(I(\mathbf{x})-c_{rb})^2 < (I(\mathbf{x})-c_{r})^2, \text{for~} \forall	r \in \{1,...,R\}, \text{and~} rb \neq r$.

Equation (\ref{eq:DNLS_MultiED}) is based on the concept of competing regions \cite{Brox2006}. Since only one level set $f_{rb}$ can be the best at each pixel, overlap of the different segments and creation of gaps are avoided without the need to add any additional coupling term.   
Only the polytopes in the immediate neighborhood of the pixel, $i \in \aleph(\mathbf{x})$, will advance or retreat to include or exclude the pixel. Therefore, the energy in equation (\ref{eq:DNLS_MultiED}) becomes minimum when the polytopes that are part of the best level set label, $rb$, include the pixel at $\mathbf{x}$, and all the remaining polytopes in the neighborhood with other labels exclude that pixel.  
For example, from Fig.~\ref{fig:MultiDemo}(d) the green and yellow polytopes retreat to exclude the pixel $\mathbf{x}$, and the green polytope advances to include the pixel. Fig.~\ref{fig:MultiDemo}(e) shows the final result with no visible overlap or gap. The energy is minimized using gradient descent by computing the derivatives of equations (\ref{eq:DNLS_MultiED}) with respect to each discriminant parameters, $w_{ijk}$.
Since we only look at a fixed number of neighboring polytopes at each pixel, the computational time of the $E_{D}$ term is independent of the number of regions $R$ to be simultaneously segmented. 

\section{Experiments}
\label{sec:Experments}
\vspace{-0.1cm}
In order to evaluate the proposed DNLS two-phase and multiphase algorithms, we compare them with their corresponding conventional signed distance function-based CV two-phase and multiphase techniques \cite{ChanVese2001, ChanVeseMultiphase}. We use the optimized implementations of the two-phase and multiphase CV level sets available in the latest Insight Segmentation and Registration Toolkit (ITK) \cite{ITKreference}. 
Our algorithm is also implemented in C++ on ITK. For quantitative comparison we use Dice coefficients (in \%) and CPU time (in seconds). 

\begin{figure}[ht!] 
\captionsetup[subfigure]{labelformat=empty}
		\begin{center}
				\subfigure{%
            \includegraphics[width=0.07\textwidth]{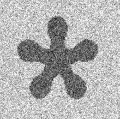}
        }%
				\subfigure{%
            \includegraphics[width=0.07\textwidth]{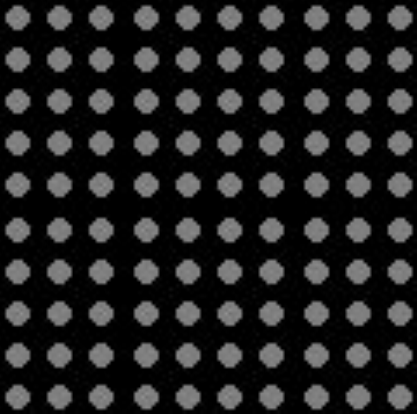}
        }%
 				\subfigure{%
           \includegraphics[width=0.07\textwidth]{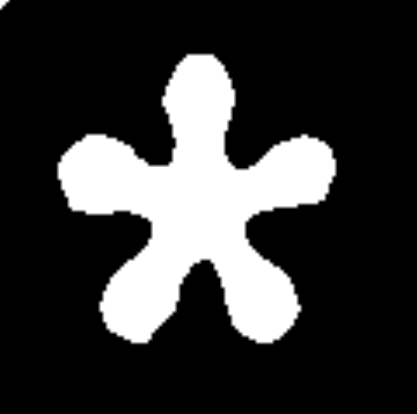}
        }%
				\subfigure{%
            \includegraphics[width=0.07\textwidth]{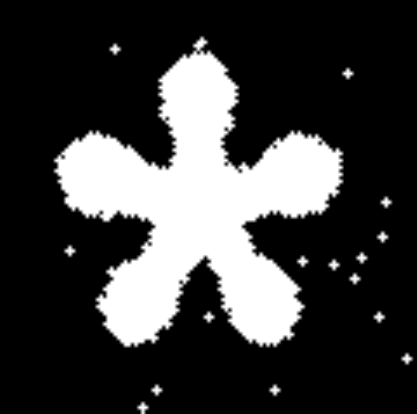}
        }%
				
								\medskip	

				\subfigure{%
            \includegraphics[width=0.07\textwidth]{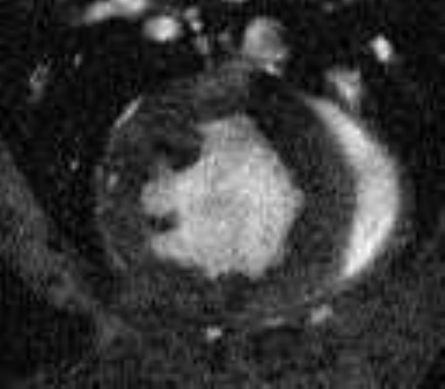}
        }%
				\subfigure{%
            \includegraphics[width=0.07\textwidth]{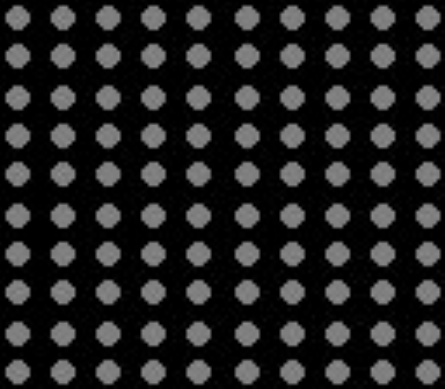}
        }%
 				\subfigure{%
           \includegraphics[width=0.07\textwidth]{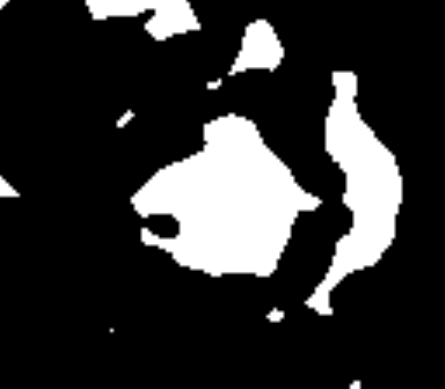}
        }%
				\subfigure{%
            \includegraphics[width=0.07\textwidth]{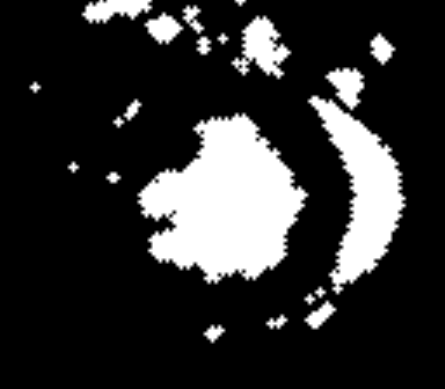}
        }
   \end{center}
   \vspace{-0.8cm}
    \caption{%
      The first and second columns show the images to be segmented and the initialization. In the third and fourth columns are the segmentation results using our DNLS and CV methods, respectively. 
     }%
   \label{fig:TwoPhase}
 \vspace{-0.4cm}
%\end{wrapfigure}
\end{figure}

\noindent
{\bf Two-phase (1 Object) Segmentation Results:}
The third and fourth columns in Fig.~\ref{fig:TwoPhase} shows the segmentation results using DNLS and CV, respectively. As can be seen from the figure, the DNLS method produces similar (or better) segmentation results compared to the CV method. However, DNLS achieves these results at a much faster computational speed. Table~\ref{table:2Phse} shows both the CPU time required and the DICE score of the segmentation results for images shown in Fig.~\ref{fig:TwoPhase}. The computational time for the CV level set is shown for both the dense and the sparse implementations available in ITK \cite{ITKreference}. It can be seen from the table that, the DNLS achieves equivalent DICE scores with a computational speed of more than 10 times compared to even the fastest sparse implementation of the CV method. DNSM parameters of $N=100$ and $M=16$ are found to be sufficient to give smooth segmentation.   
\vspace{-0.3cm}
\vspace{-0.3cm}
\begin{table}[ht!]
\caption{Quantitative comparison of DNLS and CV methods for images in Fig.~\ref{fig:TwoPhase} (phantom image top and medical bottom)}
 \vspace{-0.5cm}
\begin{center}
%\begin{scriptsize}
\begin{footnotesize}  
\begin{tabular}{|l||l|l||l|l|}
\hline
Method &\multicolumn{2}{l|}{Medical Image}&\multicolumn{2}{l|}{Phantom Image}\\
\cline{2-5}
 &DICE(\%)&TIME(s)&DICE(\%)&TIME(s)\\
\hline
CV Dense&98.2&\textbf{30.67}&98.8&\textbf{30.64}\\
CV Sparse&98.2&\textbf{5.74}&98.8&\textbf{5.67}\\
DNLS&98.4&\textbf{0.57}&99.1&\textbf{0.55}\\
\hline
\end{tabular}
%\end{scriptsize}
\end{footnotesize}
\end{center}
\label{table:2Phse}
\end{table}
\vspace{-0.4cm}

\begin{figure}[ht!] 
 \vspace{-0.1cm}
		\begin{center}
				\subfigure[]{%
            \includegraphics[width=0.058\textwidth]{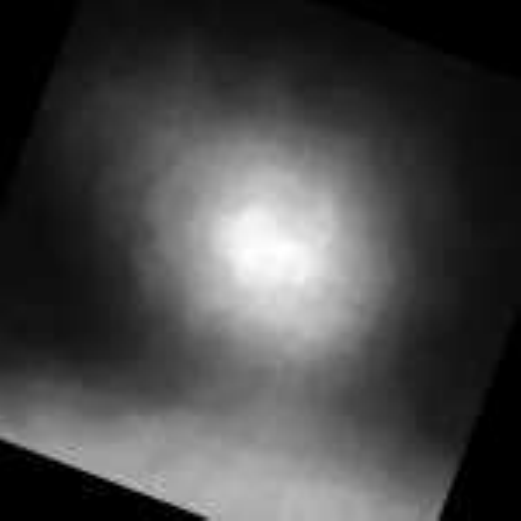}
        }%
				\subfigure[]{%
            \includegraphics[width=0.058\textwidth]{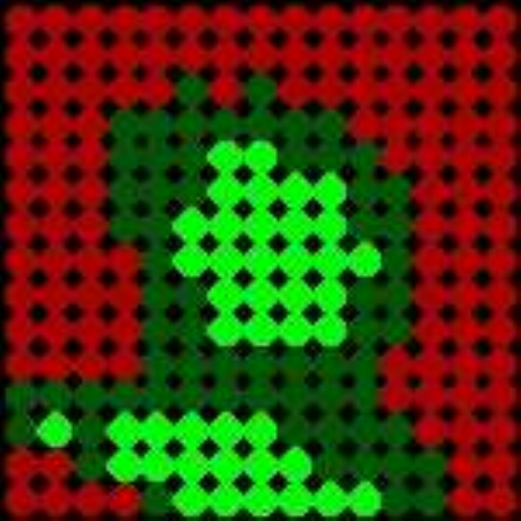}
        }%
 				\subfigure[]{%
           \includegraphics[width=0.058\textwidth]{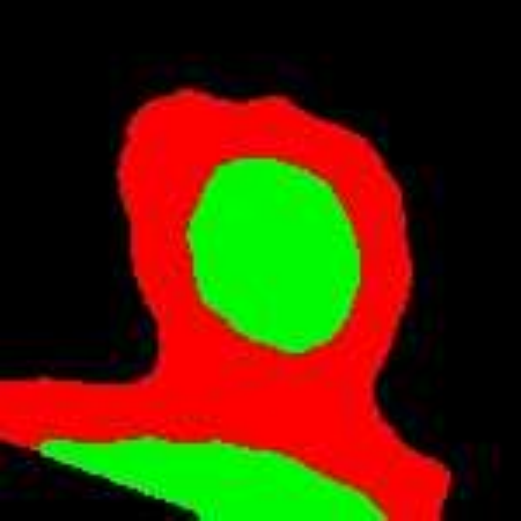}
        }%
				\subfigure[]{%
					            \includegraphics[width=0.058\textwidth]{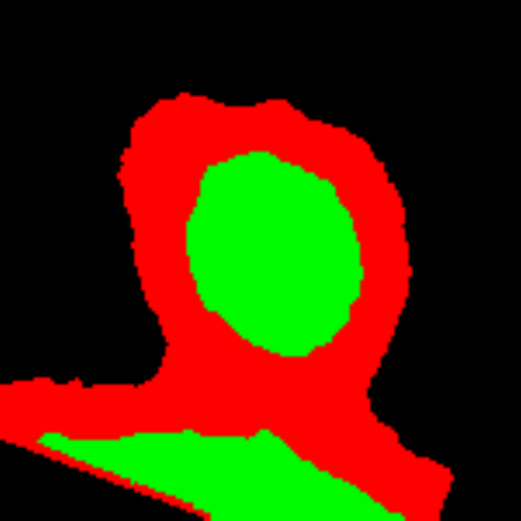}

        }%				
				\subfigure[]{%
            \includegraphics[width=0.058\textwidth]{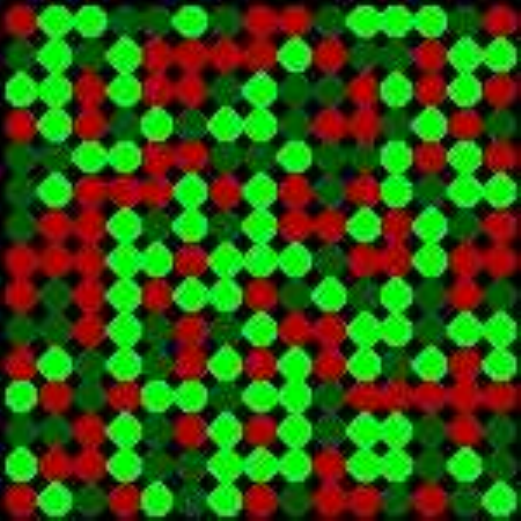}
        }%
 				\subfigure[]{%
           \includegraphics[width=0.058\textwidth]{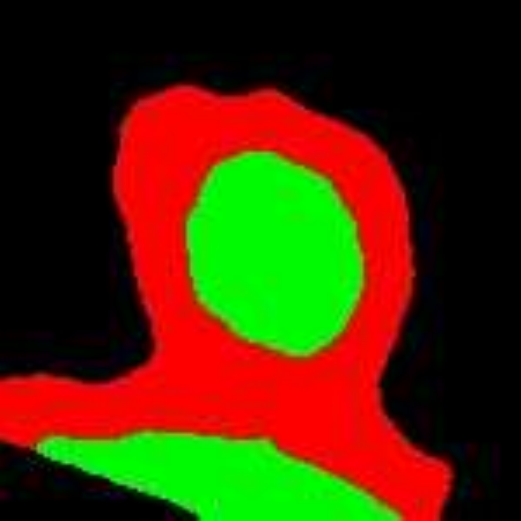}
        }%
				\subfigure[]{%
					            \includegraphics[width=0.058\textwidth]{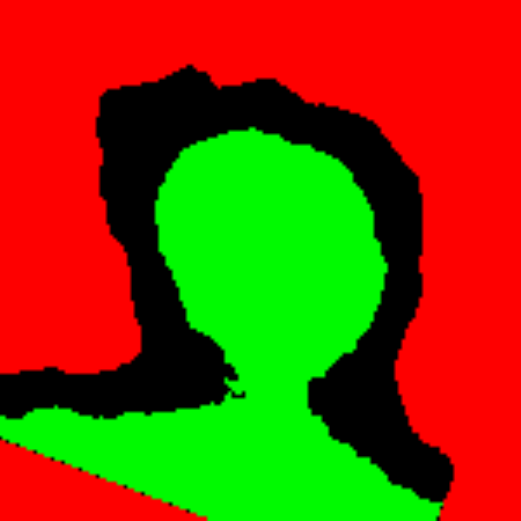}

        }%							
   \end{center}
   \vspace{-0.8cm}
    \caption{%
  %    Demonstration of the effect of initialization. 
	a) The image to be segmented. (c) and (d) are segmentations using the DNLS and CV multiphases respectively, using the Multi-Otsu threshold initialization shown in (b). (f) and (g) are segmentations using the DNLS and CV multiphases respectively, using the random initialization in (e).
     }%
   \label{fig:MultiMedical}
   \vspace{-0.3cm}
%\end{wrapfigure}
\end{figure}

\begin{figure}[ht!] 
		\begin{center}
				\subfigure[]{%
            \includegraphics[width=0.4\textwidth]{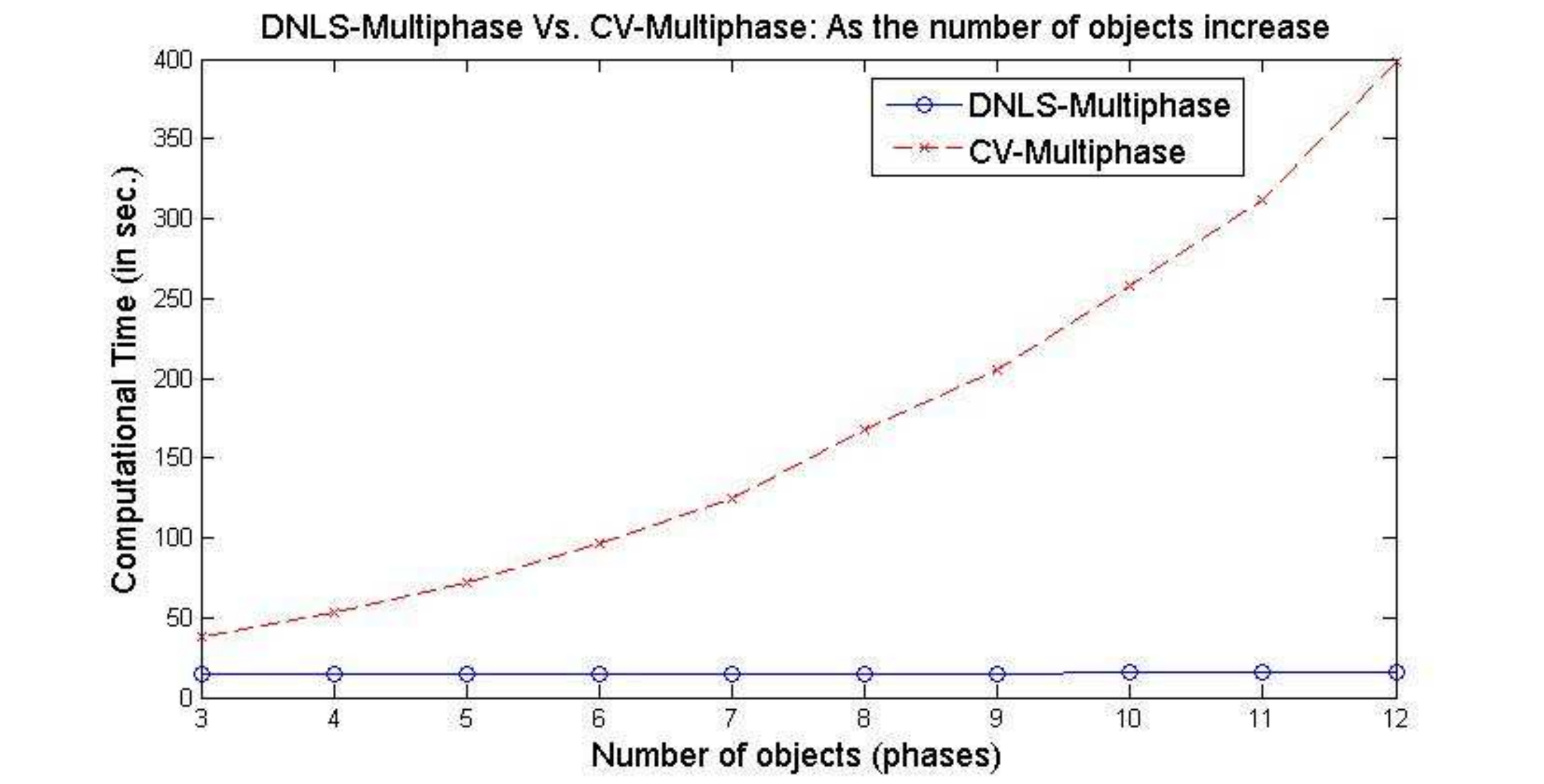}  
        }%
 				\subfigure[]{%
				 \includegraphics[width=0.08\textwidth]{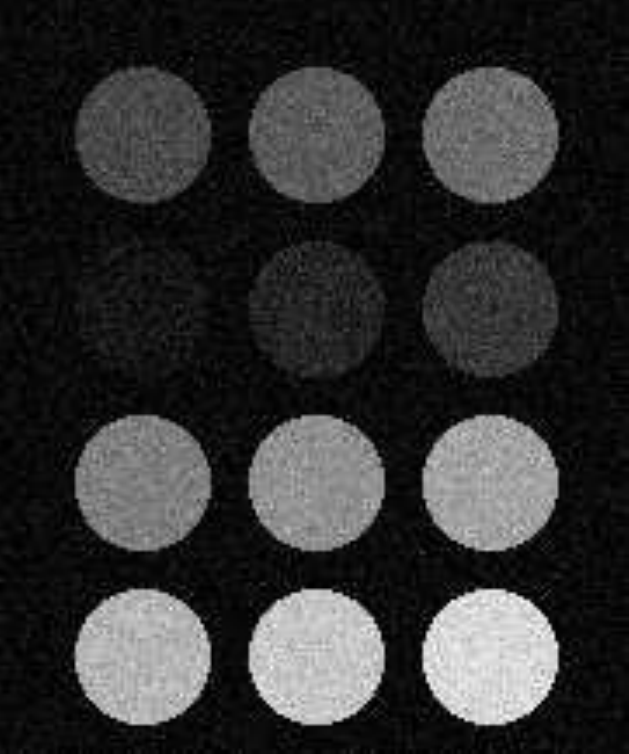}
        }%							
   \end{center}
	   \vspace{-0.8cm}
    \caption{%
      a) Computation time as a function of the number of objects. b) An example of the multi-object image used. }%
   \label{fig:ComputationalTimeVersusNumberOfPhases}
   \vspace{-0.5cm}
%\end{wrapfigure}
\end{figure}

\noindent
{\bf Multiphase Segmentation Results:}
In this section we present the results of two experiments. In the first experiment, we present the effect of initialization using 2-photon microscopy image of spine shown in Fig.~\ref{fig:MultiMedical}(a) and compare the performances of the proposed DNLS-multiphase with the standard CV-multiphase level set. 
As can be seen from Fig.~\ref{fig:MultiMedical}(c) and (d), when multi-Otsu threshold \cite{Otsu1979} is used to first obtain good initialization both the DNLS-multiphase and DNLS-multiphase give good segmentation. However, when random initialization shown in Fig.~\ref{fig:MultiMedical}(e) is used, the CV-multiphase level set converges to a poor segmentation as shown in Fig.~\ref{fig:MultiMedical}(g); whereas, our DNLS-multiphase still converges to a good segmentation as can be seen from Fig.~\ref{fig:MultiMedical}(f).  
%  \vspace{-0.3cm}
  %\vspace{-0.3cm}
In the second experiment, we show the effect of the number of objects to be segmented on the computation time. For this purpose, we generate phantom images with various number of objects; one example with 12 objects is shown in Fig.~\ref{fig:ComputationalTimeVersusNumberOfPhases}(b). 
Figure~\ref{fig:ComputationalTimeVersusNumberOfPhases} shows the computation time as a function of the number of objects in the images: using the CV-Multiphase with sparse implementation (dashed line), and using the DNLS-multiphase (solid line). The time in the graph is obtained with similar segmentation quality of around 98.5\% in DICE. The graph shows that the computation time needed by the CV-Multiphase grows exponentially as the number of objects increase. On the other hand, the proposed DNLS-Multiphase requires an almost constant computation time independent of the number of objects to be segmented. The memory required also remains constant in the proposed DNLS-multiphase, because the number of polytopes is fixed, and only their labeling changes as the number of objects change. 
% This is a crucial result specially when segmentation of images into large number of phases is required.  
  \vspace{-0.1cm}
\vspace{-0.2cm}
\section{Conclusion}
\label{sec:Conclusion}
 \vspace{-0.2cm} 
In this paper, we presented a novel parametric level set method that naturally keeps the level set function regular all the time, and that does not need any form of re-initialization or adding any regularizing term.  Due to its parametric nature the DNLS method also reduces the dimensionality of the problem and its time step is not limited by the standard CFL condition, resulting in much shorter computation time. We also presented the DNLS-multiphase framework for simultaneous segmentation of multiple objects. The proposed DNLS-multiphase approach has the highly desired properties that it is less sensitive to initialization, and its computational cost and memory requirement remain almost constant as the number of objects to be segmented grows, while also representing each object with a unique level set.

\bibliographystyle{IEEEbib}
%\bibliography{strings}
\bibliography{strings,refsICIP}

\end{document}